%% file: main.tex
\documentclass[sigconf,screen=true,anonymous=false,bookmarks=false]{acmart}
\input{setting-acm}

\usepackage{graphicx}
\graphicspath{{pdf/}}
\usepackage{enumitem}

\copyrightyear{2024}
\acmYear{2024}
\setcopyright{rightsretained}
\acmConference[GLSVLSI '24]{Great Lakes Symposium on VLSI 2024}{June
12--14, 2024}{Clearwater, FL, USA}
\acmBooktitle{Great Lakes Symposium on VLSI 2024 (GLSVLSI '24), June 12--14,
2024, Clearwater, FL, USA}
\acmDOI{10.1145/3649476.3658754}
\acmISBN{979-8-4007-0605-9/24/06}

\begin{document}

\title{
    Deep-Learning-Based Pre-Layout Parasitic Capacitance Prediction on SRAM Designs
}
\thanks{This work was partially supported by the National Science and Technology Major Project (2021ZD0114703), and NSFC under grant No. 62204141 and 62090025.
W. Yu is the corresponding author.}
\author{Shan Shen$^*$, Dingcheng Yang$^*$, Yuyang Xie$^*$, Chunyan Pei$^*$, Wenjian Yu$^*$, Bei Yu$^+$
}
\affiliation{
  \institution{*Department Computer Science \& Technology, BNRist, Tsinghua University, Beijing, China}
  \institution{$^+$Department of Computer Science \& Engineering, The Chinese University of Hong Kong, Hong Kong SAR}
}
\email{{shanshen, yu-wj}@tsinghua.edu.cn}
\input{abstract}

\maketitle
\pagestyle{plain}



\input{intro}

\input{prelim}
\input{algo}
\input{result}

\input{conclu}

{
\bibliographystyle{ACM-Reference-Format}
\bibliography{IEEEabrv, refs}
}

\end{document}

%% file: setting-acm.tex
\usepackage{enumitem}
\setlist{leftmargin=5.5mm}

\iftrue
\fi

\usepackage{lipsum}
\usepackage{graphicx}
\usepackage{amsmath}
\usepackage{footnote}
\usepackage{mathtools}
\usepackage{mathrsfs}     
\usepackage{comment}
\usepackage[subrefformat=parens,labelformat=parens]{subfig}
\usepackage{bm,bbm}
\usepackage{multirow}
\usepackage{threeparttable,booktabs}
\usepackage{blkarray}
\usepackage{tikz}
\usetikzlibrary{positioning,calc,fit,decorations.pathmorphing,shapes.geometric,shapes.gates.logic.US,calc}
\usepackage{balance}
\usepackage{courier}                                       
\usepackage{makecell}
\usepackage{cleveref}                                      
\usepackage[mathcal]{eucal}
\usepackage[]{algpseudocode}                               
\algrenewcommand\textproc{\texttt}
\makeatletter\let\float@addtolists\relax\makeatother
\usepackage{algorithm}
\usepackage{filecontents}                                  
\usepackage{pgfplots}
\usepackage{pgfplotstable}
\usepackage{pgf-pie}
\usepgfplotslibrary{groupplots}
\pgfplotsset{compat=newest}
\usepackage[figuresright]{rotating}
\usepackage{xcolor,colortbl}                               

\theoremstyle{plain}

\theoremstyle{definition}

\algrenewcommand\textproc{\texttt}

\usepackage[skip=1pt]{caption}            
\definecolor{CUHKorange}{RGB}{244,106,18} 
\definecolor{CUHKblue}{RGB}{0,111,190}    
\definecolor{CUHKgreen}{RGB}{0,127,128}   
\definecolor{CUHKred}{RGB}{228,46,36}     
\definecolor{CUHKyellow}{RGB}{198,148,34} 
\definecolor{CUHKdark}{RGB}{114,44,114}   
\definecolor{CUHKmiddle}{RGB}{144,44,144} 

\usepackage{tcolorbox}
\tcbuselibrary{skins,breakable}
    {\endtcolorbox}

%% file: abstract.tex
\begin{abstract}
To achieve higher system energy efficiency, SRAM in SoCs is often customized. The parasitic effects cause notable discrepancies between pre-layout and post-layout circuit simulations, leading to difficulty in converging design parameters and excessive design iterations. Is it possible to well predict the parasitics based on the pre-layout circuit, so as to perform parasitic-aware pre-layout simulation? In this work, we propose a deep-learning-based 2-stage model to accurately predict these parasitics in pre-layout stages. The model combines a Graph Neural Network (GNN) classifier and Multi-Layer Perceptron (MLP) regressors, effectively managing class imbalance of the net parasitics in SRAM circuits. We also employ Focal Loss to mitigate the impact of abundant internal net samples and integrate subcircuit information into the graph to abstract the hierarchical structure of schematics. Experiments on 4 real SRAM designs show that our approach not only surpasses the state-of-the-art model in parasitic prediction by a maximum of 19X reduction of error but also significantly boosts the simulation process by up to 598X speedup.

\end{abstract}


%% file: intro.tex
\section{Introduction}

With the rapid expansion of intelligent Internet of Things (IoT) devices, contemporary System-on-Chips (SoCs) are increasingly focused on enhancing energy efficiency. This is essential for extending the standby time of battery-powered devices. In order to optimize the energy efficiency of on-chip memory, both academic and industrial researchers have proposed various high-energy-efficiency memory structures. These include near-threshold SRAM  \cite{lp2}\cite{lp3}, compute-in-memory SRAM \cite{cim1}\cite{cim2}, and others.
SRAM customization involves adjustments in the topology and size of the memory cell, peripheral circuits, timing, and controller design. To ensure the stability of the chip's functionality and its final yield, it is crucial to simulate and evaluate various performance metrics, such as read/write delay, power consumption, and failure probabilities during the design procedure \cite{tcad2005}. Once the SRAM performance falls short of expectations, significant time and labor are required for iterative design modifications. This process adds complexity to customizing the SRAM IP and results in prolonged design cycles. 

In traditional design workflows, designers proceed with circuit design and optimization based on pre-layout simulations. They then perform verification using post-layout simulations after completing the layout drawing. However, with advanced technologies adopting smaller transistor sizes and lower operating voltages, there's a notable decrease in transistor driving ability. Consequently, the parasitic effect becomes too significant to be overlooked. This disparity leads to a substantial gap between pre-layout and post-layout simulation results, making it challenging to ensure the final circuit performance.
Regrettably, the back-end design of customized SRAM is an arduous and time-intensive process. This is due to two primary reasons: (1) the high density of transistors often leads to violations of design rules; (2) the routing process is complicated, as it involves managing a substantial number of data wires within a constrained area. Consequently, even minor alterations to the schematic can necessitate extensive modifications in the layout.

\begin{figure*}[t]
    \setlength{\abovecaptionskip}{0pt}
    \setlength{\belowcaptionskip}{0pt}
    \centering
    \subfloat[]{ \includegraphics[width=0.3\linewidth]{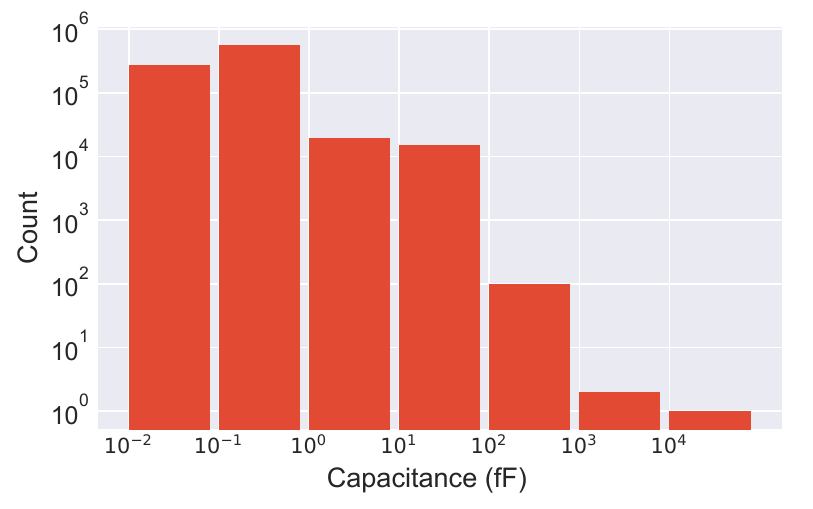} \label{fig4-a}}
    \subfloat[]{ \includegraphics[width=0.3\linewidth]{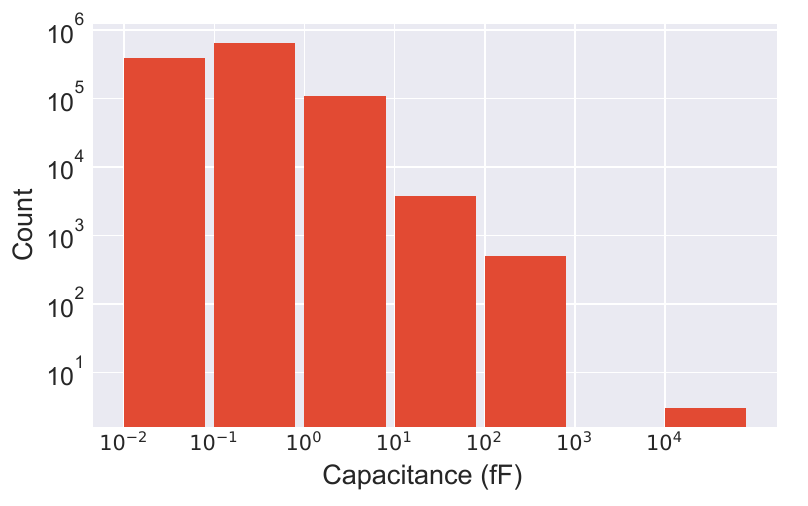} \label{fig4-b}}
    \subfloat[]{ \includegraphics[width=0.3\linewidth]{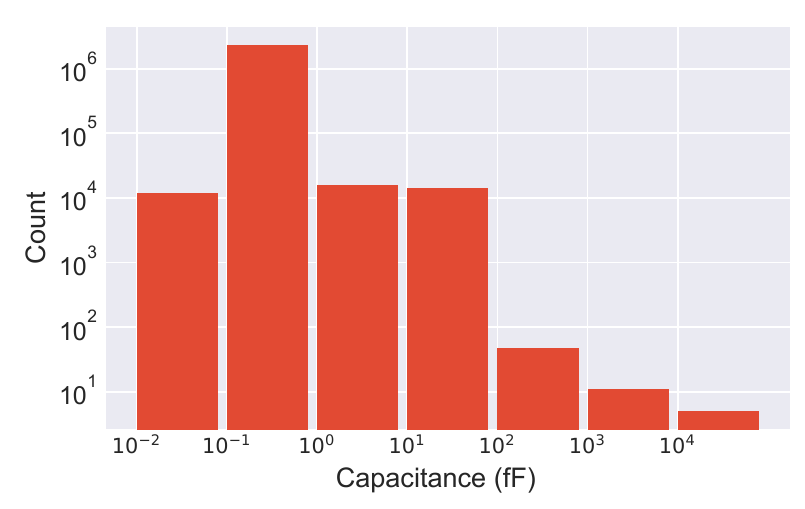} \label{fig4-c}}
    \caption{Distributions of the parasitic net capacitance of (a) Ultra8T SRAM, (b) Sandwich-RAM \cite{cim1}, and (c) SP8192W SRAM \cite{arm}.}
    \label{fig4}
\end{figure*}

Recently, numerous studies \cite{MLParest, ParaGraph, licf, date2021, 3DIC, CNNcap, icccas} have employed machine learning as a potent tool for predicting parasitic effects in electronic design. However, training accurate ML-based models is often hindered by the class imbalance of parasitic capacitance. As illustrated in \Cref{fig4}, the distribution of net capacitance shows a prominent imbalance, with values ranging from 0.01 fF to 100 pF. There are over $10^6$ nets in the second bin of SP8192W SRAM, predominantly internal connections in memory cells. This imbalance presents two problems: (1) the training process becomes inefficient, as the majority of nets are ``easy negatives'' that offer little to no valuable learning signal; (2) these easy negatives can dominate the training process, potentially leading to the development of ineffective models. In this work, we aim to address this challenge and enhance the accuracy of a deep-learning-based model for parasitic prediction. The contributions of this work are outlined as follows.
\begin{itemize}[leftmargin=0.6cm]
    \item We propose a unique 2-stage model, consisting of a Graph Neural Network (GNN) classifier and multiple Multi-Layer Perceptron (MLP) regressors, and the corresponding training strategy. 
    We implement Focal Loss \cite{fl} as the loss function of the classifier to further reduce the overwhelming effect of easy negatives. Subcircuit information is also integrated into the graph, effectively mirroring the hierarchical structure found in schematics. Our experimental results demonstrate that the 2-stage model achieves an accuracy improvement ranging from 2.5X to 19X over the state-of-the-art model \cite{ParaGraph}, while also reducing both training and inference time.
    

    \item The proposed method stands out from existing parasitic prediction models that primarily focus on small-scale analog circuits, as it targets large-scale memory circuits. By simulating the schematic netlist with the predicted parasitics, we achieve a significant speedup, up to 586X, compared to the simulations using the post-layout netlist. This approach substantially benefits the large-scale memory circuit design. The versatility of the proposed method also makes it naturally suitable for other downstream tasks. This includes design space exploration and expedited design updates. 


\end{itemize}
The paper is organized as follows.
Section \ref{relwork} reviews the related work in the field and Section \ref{prelim} introduces some fundamental concepts of GNNs.
Section \ref{algo} presents the proposed model for net capacitance prediction, and Section \ref{flow} describes the over workflow.
Section \ref{results} gives experimental results derived from real SRAM designs.
Section \ref{conclu} concludes the whole paper.

%% file: prelim.tex
\section{Related Works}\label{relwork}
In recent years, a lot of attention has been paid to machine learning as an effective method for parasitic prediction. 
Shook \textit{et al}.~\cite{MLParest} transformed the front-end netlist (schematic) of analog IP designs into a star topology and used a random forest model to predict the equivalent resistance and capacitance of each net. 
Another work, named ParaGraph \cite{ParaGraph}, converts circuit schematics into graphs and utilizes graph neural network (GNN) techniques to predict net capacitance and device layout parameters. 
It leverages a complex aggregation procedure to compute node embedding, which is a combination of graph convolutional network (GCN) \cite{gcn}, GraphSage \cite{sage}, relation GCN (RGCN) \cite{rgcn}, and graph attention network (GAT) \cite{gat}. 
It also includes the ensemble modeling technique to improve the prediction accuracy via training 3 different sub-models for different magnitudes of the capacitance value. Net samples with a ground truth larger than the maximum predicted value of the sub-model are ignored during training, which increases prediction accuracy within a specified range within each model. The sub-model with larger capacitance prediction is more preferred than that with small capacitance.
Li \textit{et al} \cite{licf} adopted a ParaGraph-like model to predict the parasitics and guide optimization of the voltage-controlled oscillator (VCO).
Liu \textit{et al}.~\cite{date2021} proposed an improved surrogate performance model using parasitic graph embeddings generated by ParaGraph \cite{ParaGraph}. Then the surrogate model was integrated into a Bayesian optimization workflow to automate transistor sizing. 

Machine learning-based capacitance extraction is also studied for the scenario of monolithic 3D (M3D) IC design. 
Pentapati \textit{et al}. \cite{3DIC} proposed a regression model based on augmented decision tree learning to better predict 3D net parasitics. 
In another scenario, Yang \textit{et al}.~\cite{CNNcap} proposed a convolutional neural network capacitance model (CNN-Cap) for the pattern-matching-based capacitance extraction. The method is able to accurately compute the capacitances of 2D patterns with a variable number of conductors.

It should be pointed out that although \cite{MLParest, ParaGraph, licf, date2021} are for predicting net capacitances at the pre-layout stage, they are all focused on analog circuits. Compared to the analog circuit, large-scale high-density memory circuits suffer from a more severe imbalance of training data.
Besides, authors in  \cite{3DIC} and \cite{icccas}, etc. assume the layout is partially ready, which is not purely based on the pre-layout schematic. In this work, we focus on predicting the net capacitances at the pre-layout stage to expedite the customized SRAM design. 

\section{Preliminaries}\label{prelim}
A graph $\mathcal{G=(V, E)}$ is a structure used to represent entities and their relations. It consists of two sets, the set of nodes $\mathcal{V}$ (also called vertices) and the set of edges $\mathcal{E}$ (also called arcs). Each node is associated with a vector of features $x_v=(x_1, ..., x_d)$ with dimension $d$. The $n=|\mathcal{V}|$ node features form a matrix $X\in{\mathbb{R}^{n\times{d}}}$. An edge $(u,v)\in\mathcal{E}$, represented as $e_{u,v}$, connecting a pair of nodes $u$ and $v$ indicates that there is a relation between them. The edge can also have a feature vector $x_e=(x_1, ..., x_c)$ with dimension $c$ and $m=|\mathcal{E}|$ features form a matrix $X^e\in{\mathbb{R}^{m\times{c}}}$.
 The neighborhood of a node $\mathcal{N}(v)$ is defined as $\mathcal N(v)=\{ u\in{\mathcal{V}}\mid{(u,v)\in\mathcal{E}} \}$. 
 Graphs can be either homogeneous or heterogeneous. In a homogeneous graph, all the nodes represent instances of the same type and all the edges represent relations of the same type. In contrast, in a heterogeneous graph, the nodes and edges can be of different types.

GNN is a kind of neural network that operates directly on data structured as graphs, without losing structural and feature information \cite{GNN}. 
Having an input graph, a GNN aims to learn the embedding vectors per node, defined as $h_u$, $\forall u\in\mathcal{V}$, which encodes the neighborhood information of each node \cite{GNNsurvey}. The message passing between nodes is assumed as the most generic GNN layer \cite{EDA}.  Given a graph structure, message passing updates the edge embeddings $h^{e}_{(u,v)}$ with
\begin{equation}
h^{e}_{(u,v)}=\phi (h_u ,h_v,x^{e}_{(u,v)}), \label{eq1}
\end{equation}
where $\phi(\cdot )$ is an arbitrary, non-linear, differentiable function that aggregates its inputs, and $x^{e}_{(u,v)}\in\mathbb{R}^{c}$ is the initial edge feature vector. After the edge embedding is obtained, and defining $x_u\in\mathbb{R}^{d}$ as the feature vector for the starting node, the node representation is updated by
\begin{equation}
h^{'}_{u}=\phi^{'} (h_u ,\sum_{v\in N(u)} h^{e}_{(u,v)},x_u). \label{eq2}
\end{equation}
The graph embeddings learned by GNNs can be used as inputs to other ML models for building an end-to-end framework. There are three levels of tasks for such a framework: node, edge, and graph \cite{GNNsurvey}. In the node-level task, the regression or classification problem of the nodes is of concern.

%% file: algo.tex
\section{Parasitic Capacitance Prediction}\label{algo}

A 2-stage deep-learning model is proposed in this section. It contains a GNN-based classifier and 5 MLP regressors. We first introduce the conversion of schematics, then the feature extraction, and last the model architecture and training strategy.

\subsection{Conversion of Circuit Netlist to Graph}\label{3A}

\begin{figure}[h]
    \setlength{\abovecaptionskip}{0pt}
    \setlength{\belowcaptionskip}{0pt}
\centering    
\includegraphics[width=0.9\linewidth]{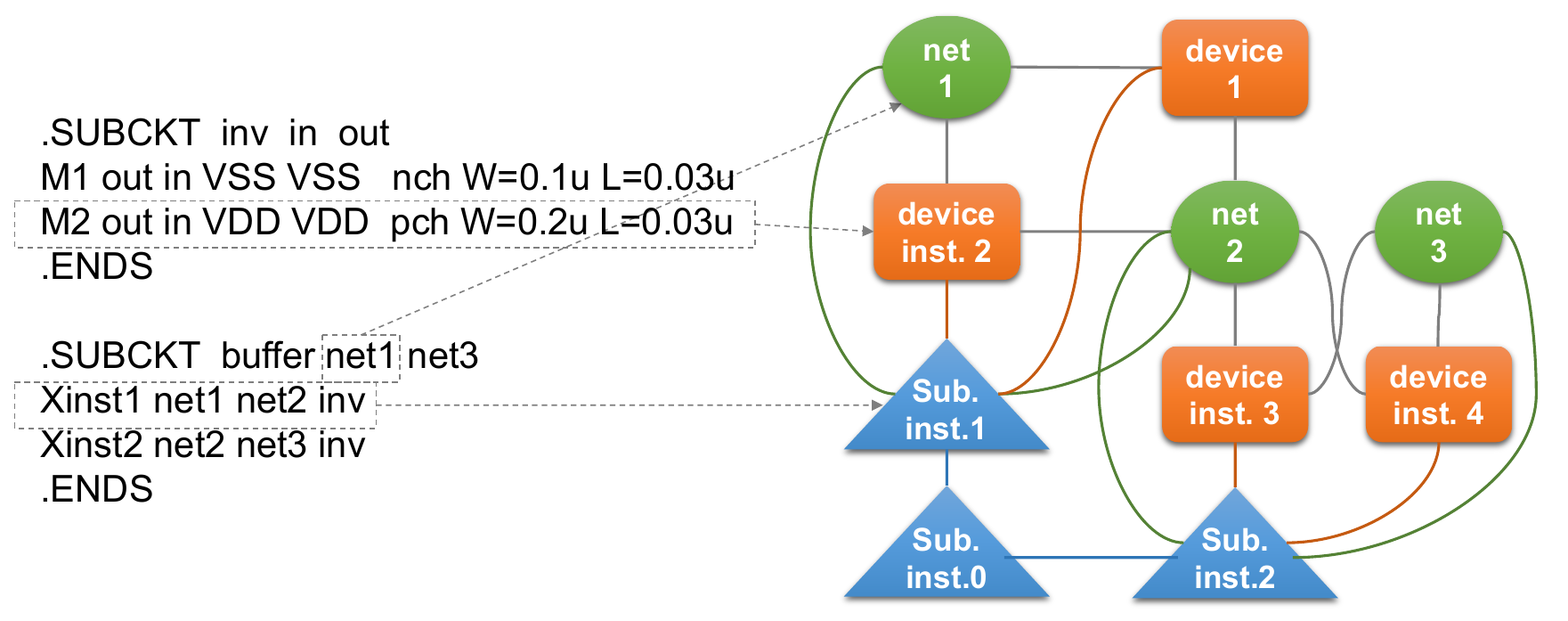}
\caption{Example of converting a circuit to a graph.}
\label{fig2}
\end{figure}

In order to reflect the modularization in circuit design, the schematic netlist will be modeled as a heterogeneous graph $\mathcal{G=(V, E)}$ in this work. A node in the graph corresponds to a net, a device, or a subcircuit instance in the circuit. 
\Cref{fig2} shows an example of a buffer containing three types of node sets $\mathcal{V}=\mathcal{V}^\mathrm{NET}\cup \mathcal{V}^\mathrm{DEV}\cup \mathcal{V}^\mathrm{SUB}$. A green circle represents a net ($v\in \mathcal{V}^\mathrm{NET}$), connecting to devices; an orange square represents an instance of the transistor device ($v\in \mathcal{V}^\mathrm{DEV}$), which can also be other types of devices such as a capacitor, a resistor, and a diode; a blue triangle represents an instance of the subcircuit ($v\in \mathcal{V}^\mathrm{SUB}$), comprised of multiple nets and device instances. We set all edges in the graph as undirected. The advantage of the undirected graph is that feature information can be transferred between different nodes in a shallow network. Compared to the graphs only comprising nodes representing nets and devices in \cite{ParaGraph}, the node of the subcircuit instance in this work can reflect the hierarchical structure in schematics and generalize the local features.

\subsection{Acquisition of Training Data}\label{3B}
In the post-layout netlist (SPF files) generated by RC extraction, the net parasitics form a complex $\pi$-type RC network. As the net capacitance is of concern, we generate the lumped sum of capacitance (C\textsubscript{eff}) for each net from the post-layout netlist and use it as the ground-truth label for training the 2-stage model.


We extract three types of feature vectors from the schematic netlist. The definitions of different feature elements are listed in \Cref{tab1}. Device nodes need to extract different features according to their device types. Compared to existing works, we collect more features for different nodes. For example, feature elements of a transistor device include the multiplier ($M_{mos}$), channel length ($L$), width ($W$), etc., while other feature elements (such as $M_{res}$, $L_{res}$, and $W_{res}$) are zeros. 
All features are normalized by the maximum value in the dataset to achieve better numerical stability. 

\begin{table}[tb!]
    \small
    \setlength{\belowcaptionskip}{0pt}
    \centering
    \caption{Feature definitions of nets, device instances, and subcircuit instances}\label{tab1}
        \begin{tabular}{|llll|}
            \hline
            Type & Notation & Definition & Index \\ 
            \hline \hline
            \multirow{13}{*}{Net} & $N_{mos}$ & \# of connected transistors & 0 \\
            & $N_{g}$ & \# of connected gate terminals & 1 \\
            & $N_{sd}$ & \# of connected source/drain terminals & 2 \\
            & $N_b$ & \# of connected base terminals & 3 \\
            & $W_{tot}$ & Total width of connected transistor & 4 \\
            & $L_{tot}$ & Total length of connected transistor & 5 \\
            & $N_{cap}$ & \# of connected capacitors & 6 \\
            & $Lr_{tot}$ & Total length of connected capacitors & 7 \\
            & $Nr_{tot}$ & Total \# of connected capacitor fingers & 8 \\
            & $N_{res}$ & \# of connected resistors & 9 \\
            & $W_{tot\_res}$ & Total width of connected resistors & 10 \\
            & $L_{tot\_res}$ & Total length of connected resistors & 11 \\
            & $N_{port}$ & \# of connected ports & 12 \\ \hline
            \multirow{11}{*}{\makecell[l]{Device \\ Instance}} & $M_{mos}$ & Multiplier of transistors & 0 \\
            & $L$ & Length of the transistor & 1 \\
            & $W$ & Width of the transistor & 2 \\
            & $M_{res}$ & Multiplier of connected resistors & 3 \\
            & $L_{res}$ & Length of resistor & 4 \\
            & $W_{res}$ & Width of resistor & 5 \\
            & $M_{cap}$ & Multiplier of connected capacitor & 6 \\
            & $Lr$ & Length of capacitor & 7 \\
            & $Nr$ & \# of capacitor fingers & 8 \\
            & $N_p$ & \# of ports in the device instance & 9 \\
            & $T$ & Type code of the device instance & 10 \\ \hline
            \multirow{4}{*}{\makecell[l]{Sub- \\ circuit \\ Instance}} & $N_{port}$ & \# of ports in the device instance & 0 \\
            & $N_d$ & \# of ports in the device instance & 1 \\
            & $N_n$ & \# of nets in the subcircuit instance & 2 \\
            & $Lvl$ & Hierarchy level of the instance & 3 \\ 
            \hline
        \end{tabular}
\end{table}

\subsection{Two-Stage Model Building}\label{3C}
The distribution of C\textsubscript{eff} is extremely imbalanced (\Cref{fig4}). The large number of nets with small capacitance, called easy negatives, results in the minority samples being prone to be mispredicted. However, those nets with large C\textsubscript{eff} are usually important ports or clock nets, which are likely to affect the timing analysis results. Therefore, we divide the training data (net nodes) into 5 categories and label them with $t\in\mathcal{T}=\{0,1,2,3,4\}$ according to the magnitude of their parasitic capacitance, i.e., \{(0.01 fF, 0.1 fF], (0.1 fF, 1 fF], (1 fF, 10 fF], (10 fF,100 fF], (100 fF,$\infty$)\}. 

\begin{figure}[!b]
    \setlength{\abovecaptionskip}{0pt}
    \setlength{\belowcaptionskip}{0pt}
    \centering
    \includegraphics[width=01\linewidth]{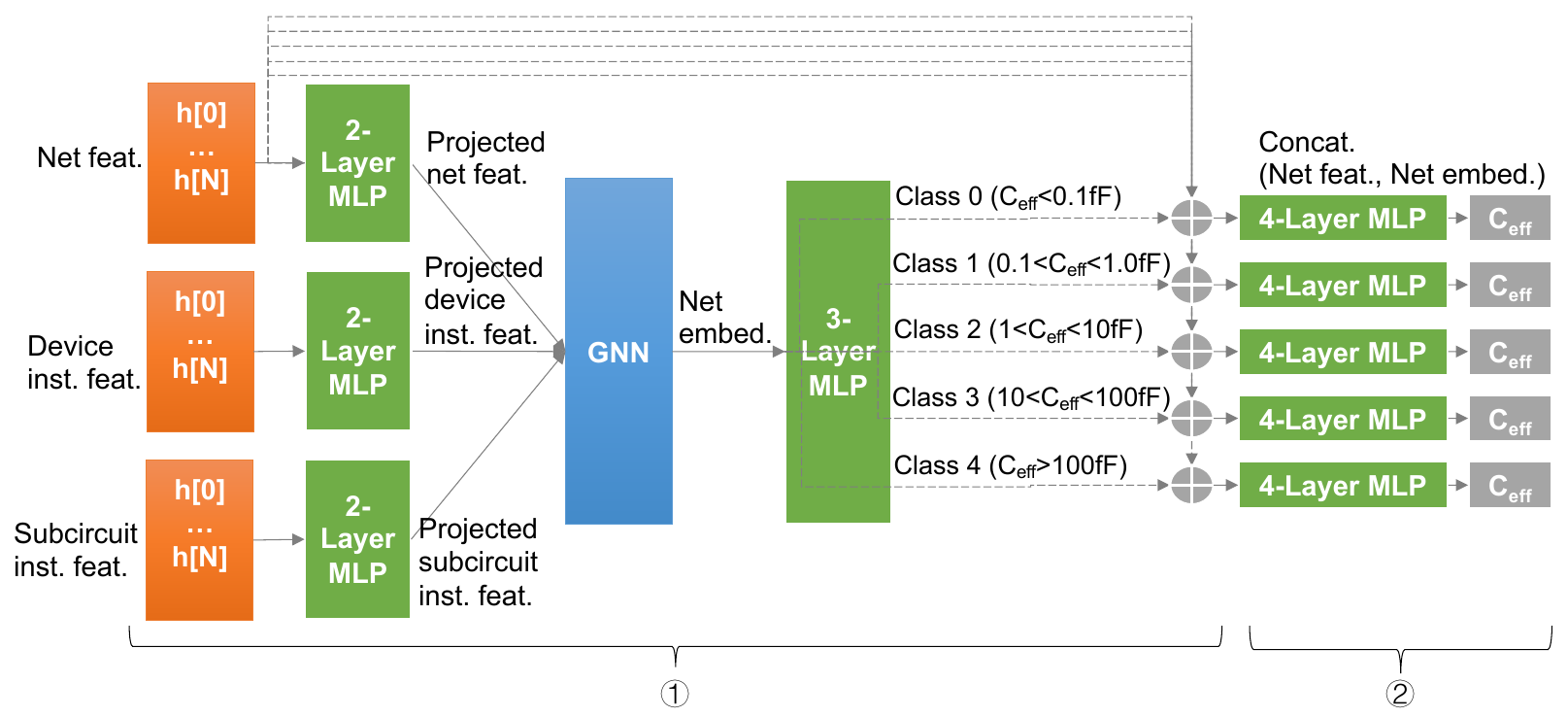}
    \caption{Structure of the proposed 2-stage model including net classification stage and capacitance regression stage.}
    \label{fig_model}
\end{figure}

GNN is more suitable for classification tasks. In order to do the capacitance regression with GNN, we build a 2-stage model based on GNN and multilayer perceptron (MLP).
The model is mainly divided into the net classification and the net capacitance regression, as shown in \Cref{fig_model}.
Feature vectors (different dimensions) of three types of nodes are projected to a common space through 3 different projectors individually so that we can easily transform a heterogeneous graph into a homogeneous one. It is convenient for subsequent training of different GNN models. Next, the projected feature vectors are input into the GNN+MLP model to classify the net nodes into different categories. Embeddings of net nodes generated by the GNN model will be fed into a 3-layer MLP to predict the label $t$ of a net node. Note that only the net nodes have labels and their embeddings are retained at this time, while that of the other two types of nodes will be masked. To reduce both the training and inference time, our proposed method leverages simple GNN models but uses a delicate training strategy and loss function.

In the 2nd stage, capacitance regression is performed for each category individually. The input of the regression model is a concatenated vector containing node embeddings and the original net feature vector. The target vector is the effective net capacitance C\textsubscript{eff}.

The training flow is also divided into 2 stages. We first use feature vectors and labels of net nodes to train and validate the classifier. Once the classifier is obtained, according to the classification result, net nodes in the training set are grouped into different sub-sets and are used to train their own regressors with the corresponding targets C\textsubscript{eff}, respectively. The test set is invisible during the whole training process.

In the 1st stage, we adopt Focal Loss \cite{fl} with a weight factor $\alpha \in [0, 1]$ which applies a modulating term to the cross-entropy loss in order to focus learning on hard examples and down-weight the numerous easy negatives.
\begin{equation}
    \mathcal{L}\!=\!\sum_{v\in{\mathcal{V}^{\mathrm{NET}}}}FL_{v}\!=\!\sum_{v\in{\mathcal{V}^{\mathrm{NET}}}}-\alpha_{t}(1-p_{v}(t))^{\gamma}\log(p_{v}(t)),
    \label{eqfl} 
\end{equation}
where $\gamma \ge 0$ is a tunable focusing parameter, $p_{v}(t)\in [0, 1]$ is the model’s estimated probability for the net node $v\in{\mathcal{V}^{\mathrm{NET}}}$ to be in class $t\in \mathcal{T}$. Here we set $\alpha$ to be 
\begin{equation}
    \alpha_{t}=\frac{1}{f_t},
\end{equation}
where ${f_t}$ is the proportion of class $t$ in the entire data samples. It can also be set to other values according to the classification results to increase the contribution of categories with fewer data to the total loss function. 

In the 2nd stage, we use the mean value of the squared percentage error as the loss function in the regression training, i.e.
\begin{equation}
    \mathcal{L}_t=\frac{1}{N_t} \sum_{v\in{\mathcal{V}^{\mathrm{NET}}}\mid t} {(\frac{y_v-\hat{y_v}}{y_v})}^2, \label{eq5} 
\end{equation}
where symbol $y$ denotes the target value and $\hat{y}$ the predicted value. $N_t$ is the number of nets with predicted label $t$.

\section{Overall Workflow}\label{flow}

\begin{figure}[tb!]
    \setlength{\abovecaptionskip}{0pt}
    \setlength{\belowcaptionskip}{0pt}
    \centering
    \includegraphics[width=0.8\linewidth]{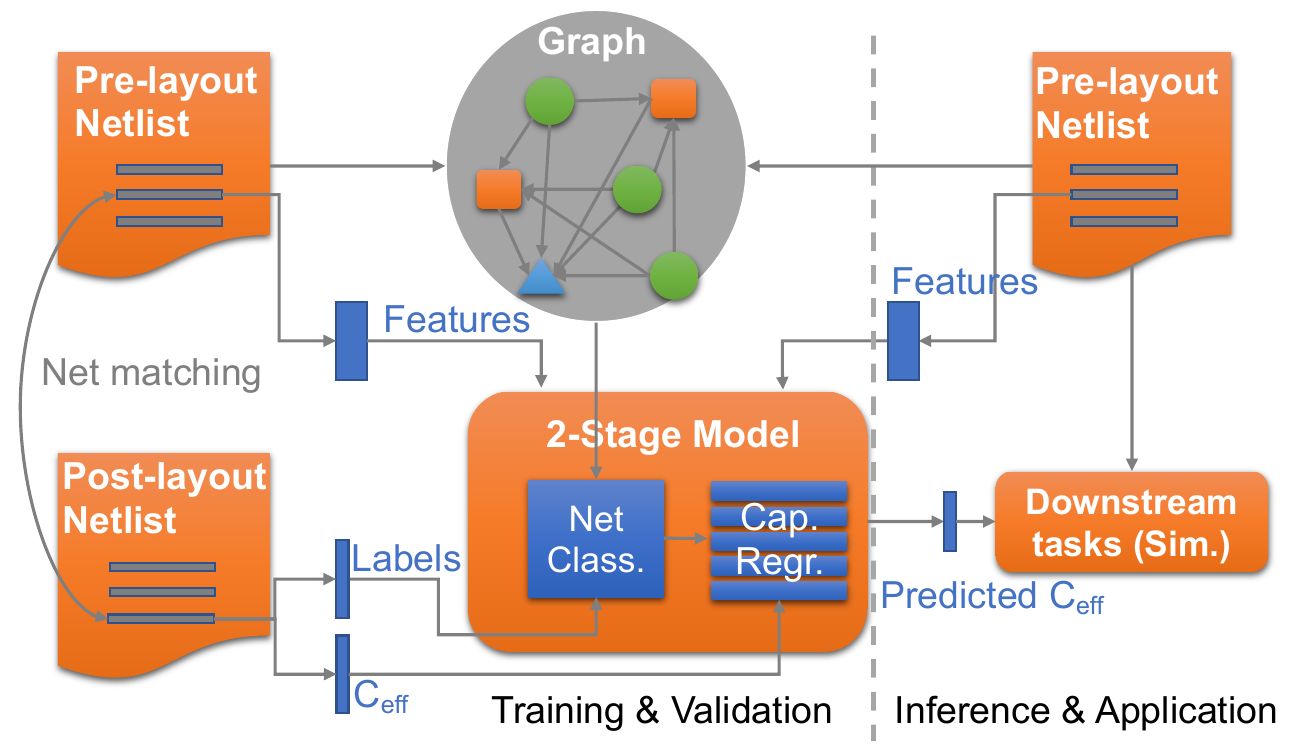}
    \caption{The overall workflow of the proposed method.}
    \label{figw}
\end{figure}

The workflow of the proposed method is illustrated in \Cref{figw}.
Training data are collected from the pre-layout schematic and the post-layout netlist by matching net names. During training, the input of the 2-stage model includes feature matrices from 3 types of nodes, a graph converted from the schematic, a class label vector, and a net capacitance vector. The output of the model is the predicted C\textsubscript{eff}. The regressors of stage 2 can be trained in a parallel manner to further speed up the model training. During the inference, the design's graph and the features of nets are fed into the model. The predicted net capacitance from the model is back-annotated into a netlist for downstream tasks.

%% file: result.tex
\section{Experimental Results}\label{results}

Experiments are conducted to show the effectiveness of the proposed method with 4 SRAM designs. All ML models are implemented based on DGL library \cite{dgl} written in PyTorch. The post-layout netlists with full parasitics are extracted by StarRC.
All experiments are run on a server with 40 Intel Xeon Silver CPUs with 128GB memory.

\begin{table}[h]
    \small
    \setlength{\abovecaptionskip}{0pt}
    \setlength{\belowcaptionskip}{0pt}
    \centering
    \caption{SRAM designs used by this work.}\label{tab2}
    \resizebox{\linewidth}{!}{
    \begin{tabular}{|l|cccc|}
        \hline
        SRAM Designs & SSRAM & Ultra8T & Sandwich & SP8192W \\ \hline \hline
        \# of Nets & 19902 & 861842 & 1160940 & 2342588 \\
        \# of Device Inst. & 57417 & 2325092 & 2665422 & 6984821 \\
        \# of Subcircuit Inst. & 9965 & 315466 & 428498 & 39885 \\
        Total \# of Nodes & 87284 & 3502400 & 4254860 & 9367294 \\
        \# of Edges & 134926 & 13392268 & 13254854 & 32009072 \\ \hline
    \end{tabular}}
\end{table}

\subsection{Dataset}\label{4A}
We prepare the full schematic netlists and the post-layout SPF netlists to extract the feature vectors and net capacitance respectively.
\Cref{tab2} summarizes the SRAM design examples utilized by this work (net parasitic capacitance distributions are illustrated in \Cref{fig4}).
All designs are under 28nm CMOS technology. SSRAM \cite{lp2} is a small design with high-energy efficiency with a timing-speculation technique. The Sandwich-RAM \cite{cim1} is made up of half of digital circuits for computing and half of memory arrays, forming a sandwich-like structure. Ultra8T SRAM \cite{ultra8t} is a multi-voltage design with a wide range of operation voltages, which contains analog circuits. SP8192W SRAM is based on a single port 6T cell structure generated by the SRAM compiler \cite{arm} with tremendously high density. We leverage the full neighbor sampling method provided by DGL, which randomly splits the test set from the original large graph and returns a new subgraph. This ensures the invisibility of nodes in the test set during model training. The train/validation/test set split ratio for each design is 0.6/0.2/0.2.

\subsection{Model Settings}\label{4B}
In the 1st stage, we adopt three mainstream GNN structures, including a 3-layer graph attention network (GAT) \cite{gat}, a 3-layer graph convolutional networks (GCN)~\cite{gcn}, and 2-layer GraphSAGE \cite{sage}, in net classification.
GraphSAGE can be implemented with different types of aggregators in \eqref{eq1} and \eqref{eq2}, and we use `mean' and `pool' in our comparison.
We set all layer widths to 64. The feature projection MLP has 2 linear layers and the classification MLP has 3 linear layers. The activation function is ReLU and the dropout is 0.1 for all layers. The batch normalization (BN) is turned on. The GAT-based classifier needs to turn on the layer normalization and turn off the BN to get reasonable accuracy. We use a cosine annealing schedule to adjust the learning rate based on the number of executed epochs. The learning rate ranges from 1E-3 to 1E-4. The weight decay parameter is set to 5E-4. The performance metrics include accuracy, and the F1 macro score (the best value is 1.0 for all metrics). F1 macro is the unweighted mean of the F1 score of each class. The training process is run 10 times for each model and the performance metrics are averaged from the best epochs.

In the 2nd stage, each regressor is a 4-layer MLP with hidden layer widths \{128, 128, 64\}. The input width is the sum of the node embedding width 64 and the original net feature width 13 while the output width is 1. The activation function is ReLU and the dropout is 0.5. The learning rate is set to 1E-3 and the weight decay is set to 5E-4. The following mean absolute percentage error (MAPE)  is used as the performance metric:
\begin{equation}
    \mathrm{MAPE}=\frac{100\%}{N} \sum_{i}^{N} {\left |\frac{y_i-\hat{y_i}}{y_i}\right |}. \label{eq6} 
\end{equation}

We also implement ParaGraph \cite{ParaGraph}. ParaGraph is comprised of 3 sub-models to predict the net capacitance ranging from (0.01fF, 1fF], (1fF, 10fF], (10fF, 10pF], respectively. Each submodel has a 32-dimension width, a 5-layer GNN model, and a 4-layer MLP regressor.

\subsection{Classification and Regression Results}\label{4C}
\Cref{tab4} shows the classification results using different GNN models.
The GraphSAGE-based classifier has the best accuracy across all design cases. The attention-based classifier has relatively lower F1 macro scores than others. Besides, SP8192W SRAM has special characteristics since all classifiers have reduced scores in this case.
It can be explained by the imbalanced C\textsubscript{eff} distribution in \Cref{fig4}, where the number of easy negatives from class 1 is over 100X larger than that of other classes.

\begin{figure}[!b]
    \centering
\includegraphics[width=0.9\linewidth]{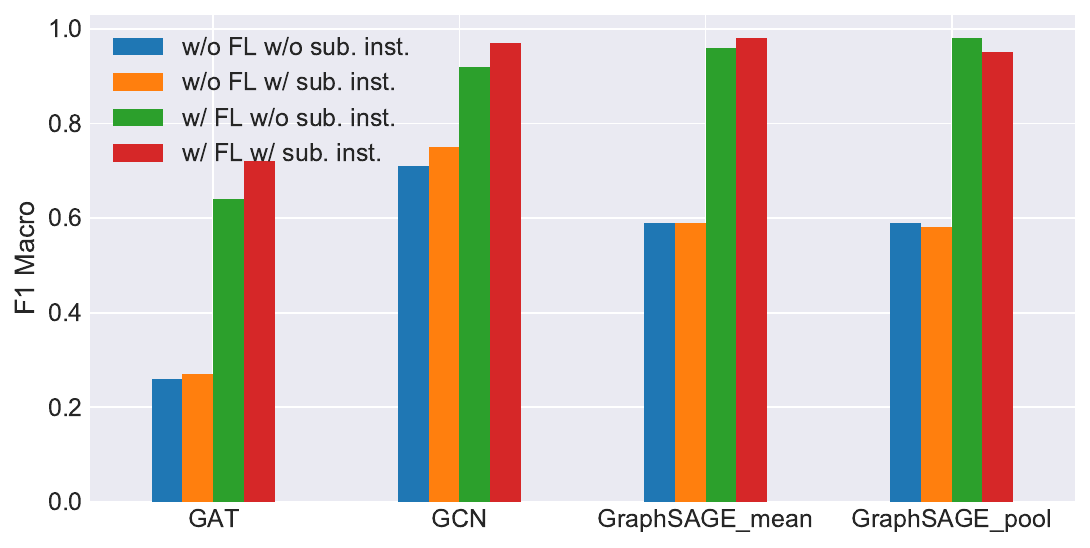}
    \caption{Performance comparison of the GNN-based classifiers trained by different strategies.}
    \label{fig6}
\end{figure}

In \Cref{fig6}, we compare the classification results with different 
training strategies. After changing the loss function from cross entropy to Focal loss, the F1 macro scores are improved significantly. Moreover, by introducing the subcircuit instance nodes in the graph, the hierarchical information improves the performance of the GAT-based, GCN, GraphSAGE\_mean-based classifiers, but slightly degenerates that of the GraphSAGE\_pool-based model. We also find the existence of subcircuit nodes helps GCN and GraphSAGE\_mean to converge more quickly during training.

\begin{table*}[tb!]
    \footnotesize
    \centering
    \caption{The classification accuracy, F1 macro, and mean absolute percentage error (MAPE) of the regressors with different GNN classifiers on the test set. (The figures meaning the best performance are highlighted)}\label{tab4}
    \begin{tabular}{|lc|cc|cccccc|}
        \hline
        \multicolumn{1}{|c}{Design Case} &\multicolumn{1}{|c|}{Choice of Classifier} & Class Acc. & F1 Macro & Class 0 & Class 1 & Class 2 & Class 3 & Class 4 & All tested nets \\ \hline \hline
        \multicolumn{1}{|l|}{\multirow{5}{*}{SSRAM}}  & ParaGraph\cite{ParaGraph} & - & - & - & - & - & -  & - & 33.53\% \\ 
        \multicolumn{1}{|l|}{} & None & - & - & 21.73\% & 5.53\% & 28.84\% & 6.59\% & - & 6.95\% \\
        \multicolumn{1}{|l|}{} & GAT & 93.34\% & 0.83 & 26.15\% & \textbf{0.88\%} & 26.37\% & 1.68\% & - & 4.30\% \\ 
        \multicolumn{1}{|l|}{} & GCN & 95.38\% & 0.88 & 15.58\% & 2.22\% & 19.00\% & 5.66\% & - & 4.05\% \\
        \multicolumn{1}{|l|}{} & GraphSAGE\_mean & 97.98\% & 0.93 & 24.19\% & 1.37\% & \textbf{14.09\%} & 6.35\% & - & \textbf{3.13\%} \\
        \multicolumn{1}{|l|}{} & GraphSAGE\_pool & \textbf{99.22}\% & \textbf{0.97} & \textbf{14.11\%} & 8.70\% & 31.94\% & \textbf{1.34\%} & - & 9.66\% \\ \hline
        \multicolumn{1}{|l|}{\multirow{6}{*}{Ultra8T}} & ParaGraph\cite{ParaGraph} & - & - & - & - & - & -  & - & 32.01\% \\ 
        \multicolumn{1}{|l|}{} & None & - & - & 4.12\% & 11.59\% & 28.81\% & 20.79\% & 9.49\% & 9.78\% \\
        \multicolumn{1}{|l|}{} & GAT & 96.61\% & 0.90 & 2.57\% & \textbf{1.30\%} & 17.10\% & 5.25\% & 45.07\% & 2.67\% \\ 
        \multicolumn{1}{|l|}{} & GCN & 98.36\% & 0.94 & 1.15\% & 1.59\% & \textbf{6.11\%} & 4.89\% & 9.10\% & \textbf{1.68\%} \\
        \multicolumn{1}{|l|}{} & GraphSAGE\_mean & \textbf{99.91\%} & \textbf{0.99} & \textbf{0.95\%} & 1.87\% & 6.77\% & \textbf{3.56\%} & \textbf{5.63\%} & 1.72\% \\
        \multicolumn{1}{|l|}{} & GraphSAGE\_pool & 98.29\% & 0.96 & 3.10\% & 9.89\% & 14.07\% & 5.92\% & 10.03\% & 7.74\% \\ \hline
        \multicolumn{1}{|l|}{\multirow{6}{*}{Sandwich}}  & ParaGraph\cite{ParaGraph} & - & - & - & - & - & -  & - & 34.98\% \\ 
        \multicolumn{1}{|l|}{} & None & - & - & 23.57\% & 25.05\% & 29.01\% & 55.78\% & 8.20\% & 25.07\% \\
        \multicolumn{1}{|l|}{} & GAT & 87.04\% & 0.72 & 15.92\% & 10.08\% & 20.43\% & 30.08\% & 35.28\% & 13.60\% \\
        \multicolumn{1}{|l|}{} & GCN & 97.99\% & 0.97 & 3.83\% & 6.98\% & 10.12\% & 10.46\% & \textbf{6.71\%} & 6.25\% \\
        \multicolumn{1}{|l|}{} & GraphSAGE\_mean & \textbf{98.72\%} & \textbf{0.98} & \textbf{3.11}\% & \textbf{5.96\%} & \textbf{8.56\%} & \textbf{8.21\%} & 8.49\% & \textbf{5.32\%} \\
        \multicolumn{1}{|l|}{} & GraphSAGE\_pool & 96.33\% & 0.95 & 3.84\% & 7.91\% & 10.99\% & 14.07\% & 8.06\% & 6.83\% \\ \hline
        \multicolumn{1}{|l|}{\multirow{5}{*}{SP8192W}} & ParaGraph\cite{ParaGraph} & - & - & - & - & - & -  & - & 4.61\% \\ 
        \multicolumn{1}{|l|}{} & None & - & - & 26.68\% & 1.14\% & 38.27\% & 6.74\% & 55.75\% & 1.62\% \\
        \multicolumn{1}{|l|}{} & GAT & 99.17\% & 0.72 & 33.34\% & 0.36\% & 25.95\% & 2.71\% & 9.34\% & 0.87\% \\ 
        \multicolumn{1}{|l|}{} & GCN & 99.22\% & 0.76 & 26.32\% & \textbf{0.26\%} & 22.70\% & 2.74\% & 21.18\% & \textbf{0.81\%} \\
        \multicolumn{1}{|l|}{} & GraphSAGE\_mean & \textbf{99.72\%} & \textbf{0.88} & \textbf{20.59\%} & 0.73\% & \textbf{12.36\%} & \textbf{2.34\%} & \textbf{3.99\%} & 1.04\% \\
        \multicolumn{1}{|l|}{} & GraphSAGE\_pool & 99.40\% & 0.82 & 24.38\% & 1.31\% & 26.66\% & 5.33\% & 6.10\% & 1.82\% \\ \hline
    \end{tabular}
\end{table*}

\Cref{tab4} also compares the accuracy of different ML-based regression models. ParaGraph's prediction errors are listed in the last column, as it does not incorporate the net classification stage. ParaGraph has the worst prediction accuracy, over 30\% MAPEs for the first 3 design cases. This is because it fails to solve the class imbalance in the dataset.
The `None' row from each design case represents the 5 pure regression models without introducing any classification and feature projection. They are trained individually according to the ground truth labels of samples. The input of the regressors is just the original features of net nodes. Compared to the `None' model, GNN-based models have lower predicted errors across all net classes.
For the GraphSAGE model, using a `mean' aggregator is a good choice in our experiments and achieves better accuracy. The GCN-base and the GraphSAGE\_mean-based regression models have similar accuracy across all design cases. The minimum and maximum MAPE reductions of the proposed 2-stage models against ParaGraph are 2.5X in SP8192W and 19X in Ultra8T.
Notice that for SP8192W SRAM, the MAPE increases for all models when predicting C\textsubscript{eff} of the nets in class 0 and class 2. This is due to a lot of false positives residing in these 2 classes (see the low F1 macro scores) as noise.

\subsection{Other Comparisons}\label{4D}
\begin{figure}[!b]
    \centering
    \includegraphics[width=0.96\linewidth]{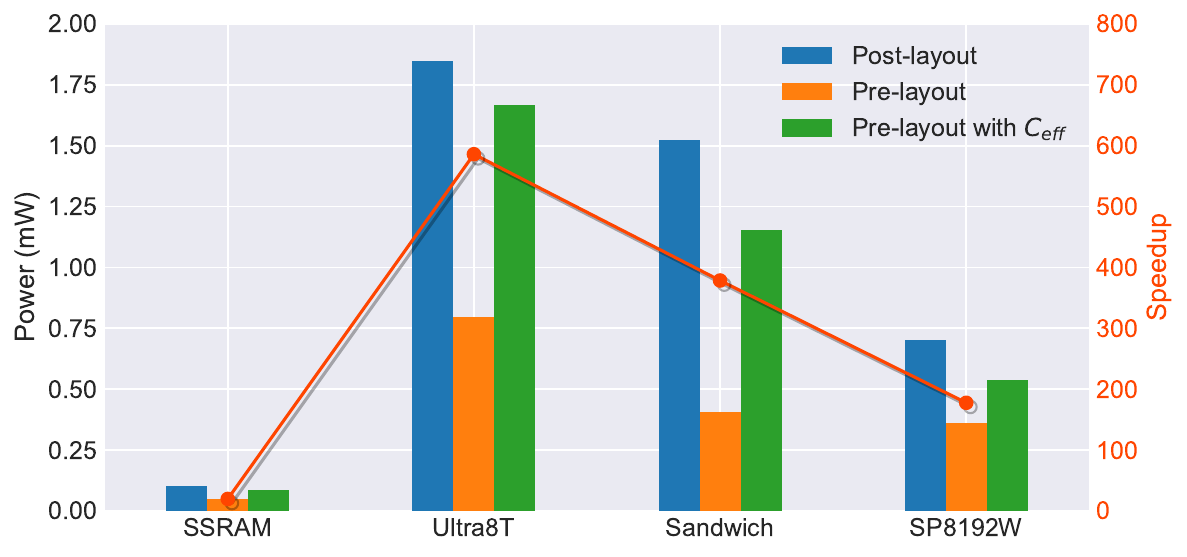}
    \caption{Simulated power consumption using different netlists, and the speedup of the proposed method.}
    \label{fig9}
\end{figure}

\Cref{fig9} shows the power consumption collected from the pre-layout simulation, the post-layout simulation, and the proposed method. With the predicted C\textsubscript{eff}, the performance error of the pre-layout simulation is largely reduced from 57.24\% to 17.77\% on average. Moreover, since our method only uses schematics, the maximum simulation speedup reaches 586X for Ultra8T while the minimum speedup is 19.66X for SSRAM compared to the post-layout simulation.

\Cref{tab5} further lists the scales of different models, and the training/inference time. ParaGraph \cite{ParaGraph} exhibits the largest training and inference time due to the complicated aggregation function that requires an attention operation for each edge type. In general, the proposed GraphSAGE\_mean-based model is the most efficient model. The proposed models have a larger number of trainable parameters than ParaGraph due to the existence of the 5 regressors.


\begin{table}[!h]
    \begin{center}
    \caption{Storage and time overhead of different GNN-based models.}\label{tab5}
    \resizebox{\linewidth}{!}{
    \begin{tabular}{|l|ccccc|}
    \hline
    Info. & ParaGraph\cite{ParaGraph} & GAT & GCN & GraphSAGE\_mean & GraphSAGE\_pool \\ \hline \hline
    \# of params & 141,987 & 212,362 & 212,618 & 216,650 & 224,970 \\
    train. time(h) & 21.59 & 16.43 & 13.41 & 7.65 & 15.94 \\
    infer. time(s) & 27.10 & 17.28 & 5.58 & 4.07 & 6.08 \\ \hline
    \end{tabular}}
    \end{center}
\end{table}

%% file: conclu.tex
\section{Conclusion}\label{conclu}
This paper presents a novel method to train a 2-stage model based on GNN and MLP  for predicting parasitic capacitances in SRAM designs. This model well handles the class imbalance of net parasitics in SRAMs, and thus outperforms the existing state-of-the-art model.
In the future, the proposed method will be extended to complete RC prediction and integrated into circuit optimization algorithms of energy-efficient SRAM design. 




%% file: main.bbl

\begin{thebibliography}{23}


\ifx \showCODEN    \undefined \def \showCODEN     #1{\unskip}     \fi
\ifx \showDOI      \undefined \def \showDOI       #1{#1}\fi
\ifx \showISBNx    \undefined \def \showISBNx     #1{\unskip}     \fi
\ifx \showISBNxiii \undefined \def \showISBNxiii  #1{\unskip}     \fi
\ifx \showISSN     \undefined \def \showISSN      #1{\unskip}     \fi
\ifx \showLCCN     \undefined \def \showLCCN      #1{\unskip}     \fi
\ifx \shownote     \undefined \def \shownote      #1{#1}          \fi
\ifx \showarticletitle \undefined \def \showarticletitle #1{#1}   \fi
\ifx \showURL      \undefined \def \showURL       {\relax}        \fi
\providecommand\bibfield[2]{#2}
\providecommand\bibinfo[2]{#2}
\providecommand\natexlab[1]{#1}
\providecommand\showeprint[2][]{arXiv:#2}

\bibitem[ARM(2023)]%
        {arm}
\bibfield{author}{\bibinfo{person}{ARM}.} \bibinfo{year}{2023}\natexlab{}.
\newblock \bibinfo{title}{{Artisan Embedded Memory IP}}.
\newblock
\newblock
\urldef\tempurl%
\url{https://www.arm.com/en/products/silicon-ip-physical/embedded-memory}
\showURL{%
\tempurl}


\bibitem[Chien and Wang(2018)]%
        {lp3}
\bibfield{author}{\bibinfo{person}{Yung-Chen Chien} {and} \bibinfo{person}{Jinn-Shyan Wang}.} \bibinfo{year}{2018}\natexlab{}.
\newblock \showarticletitle{{A 0.2 V 32-Kb 10T SRAM with 41 nW standby power for IoT applications}}.
\newblock \bibinfo{journal}{\emph{{IEEE} Trans. Circuits Syst. {I}}} \bibinfo{volume}{65}, \bibinfo{number}{8} (\bibinfo{year}{2018}), \bibinfo{pages}{2443--2454}.
\newblock


\bibitem[Gong et~al\mbox{.}(2010)]%
        {icccas}
\bibfield{author}{\bibinfo{person}{Weibing Gong}, \bibinfo{person}{Wenjian Yu}, \bibinfo{person}{Yongqiang L{\"u}}, \bibinfo{person}{Qiming Tang}, \bibinfo{person}{Qiang Zhou}, {and} \bibinfo{person}{Yici Cai}.} \bibinfo{year}{2010}\natexlab{}.
\newblock \showarticletitle{A parasitic extraction method of {VLSI} interconnects for pre-route timing analysis}. In \bibinfo{booktitle}{\emph{Proc. Int. Conf. on Commun., Circuits and Syst. (ICCCAS)}}. \bibinfo{pages}{871--875}.
\newblock


\bibitem[Gori et~al\mbox{.}(2005)]%
        {GNN}
\bibfield{author}{\bibinfo{person}{Marco Gori}, \bibinfo{person}{Gabriele Monfardini}, {and} \bibinfo{person}{Franco Scarselli}.} \bibinfo{year}{2005}\natexlab{}.
\newblock \showarticletitle{A new model for learning in graph domains}. In \bibinfo{booktitle}{\emph{Proc. Int. Joint Conf. on Neural Netw. (IJCNN)}}. \bibinfo{pages}{729--734}.
\newblock


\bibitem[Hamilton et~al\mbox{.}(2017)]%
        {sage}
\bibfield{author}{\bibinfo{person}{Will Hamilton}, \bibinfo{person}{Zhitao Ying}, {and} \bibinfo{person}{Jure Leskovec}.} \bibinfo{year}{2017}\natexlab{}.
\newblock \showarticletitle{Inductive representation learning on large graphs}.
\newblock \bibinfo{journal}{\emph{Advances in Neural Information Process. Syst.}}  \bibinfo{volume}{30} (\bibinfo{year}{2017}).
\newblock


\bibitem[Kipf and Welling(2016)]%
        {gcn}
\bibfield{author}{\bibinfo{person}{Thomas~N Kipf} {and} \bibinfo{person}{Max Welling}.} \bibinfo{year}{2016}\natexlab{}.
\newblock \showarticletitle{Semi-supervised classification with graph convolutional networks}.
\newblock \bibinfo{journal}{\emph{arXiv preprint arXiv:1609.02907}} (\bibinfo{year}{2016}).
\newblock


\bibitem[Li et~al\mbox{.}(2023)]%
        {licf}
\bibfield{author}{\bibinfo{person}{Chenfeng Li}, \bibinfo{person}{Dezhong Hu}, {and} \bibinfo{person}{Xiaoyan Zhang}.} \bibinfo{year}{2023}\natexlab{}.
\newblock \showarticletitle{Pre-Layout Parasitic-Aware Design Optimizing for RF Circuits Using Graph Neural Network}.
\newblock \bibinfo{journal}{\emph{Electronics}} \bibinfo{volume}{12}, \bibinfo{number}{2} (\bibinfo{year}{2023}), \bibinfo{pages}{465}.
\newblock


\bibitem[Lin et~al\mbox{.}(2017)]%
        {fl}
\bibfield{author}{\bibinfo{person}{Tsung-Yi Lin}, \bibinfo{person}{Priya Goyal}, \bibinfo{person}{Ross Girshick}, \bibinfo{person}{Kaiming He}, {and} \bibinfo{person}{Piotr Doll{\'a}r}.} \bibinfo{year}{2017}\natexlab{}.
\newblock \showarticletitle{Focal loss for dense object detection}. In \bibinfo{booktitle}{\emph{Proc. Int. Conf. on Comput. Vision (ICCV)}}. \bibinfo{pages}{2980--2988}.
\newblock


\bibitem[Liu et~al\mbox{.}(2021)]%
        {date2021}
\bibfield{author}{\bibinfo{person}{Mingjie Liu}, \bibinfo{person}{Walker~J Turner}, \bibinfo{person}{George~F Kokai}, \bibinfo{person}{Brucek Khailany}, \bibinfo{person}{David~Z Pan}, {and} \bibinfo{person}{Haoxing Ren}.} \bibinfo{year}{2021}\natexlab{}.
\newblock \showarticletitle{Parasitic-aware analog circuit sizing with graph neural networks and {Bayesian} optimization}. In \bibinfo{booktitle}{\emph{Proc. DATE}}. \bibinfo{pages}{1372--1377}.
\newblock


\bibitem[Lopera et~al\mbox{.}(2021)]%
        {EDA}
\bibfield{author}{\bibinfo{person}{Daniela~S{\'a}nchez Lopera}, \bibinfo{person}{Lorenzo Servadei}, \bibinfo{person}{Gamze~Naz Kiprit}, \bibinfo{person}{Souvik Hazra}, \bibinfo{person}{Robert Wille}, {and} \bibinfo{person}{Wolfgang Ecker}.} \bibinfo{year}{2021}\natexlab{}.
\newblock \showarticletitle{A survey of graph neural networks for electronic design automation}. In \bibinfo{booktitle}{\emph{Proc. MLCAD}}. \bibinfo{pages}{1--6}.
\newblock


\bibitem[Mukhopadhyay et~al\mbox{.}(2005)]%
        {tcad2005}
\bibfield{author}{\bibinfo{person}{Saibal Mukhopadhyay}, \bibinfo{person}{Hamid Mahmoodi}, {and} \bibinfo{person}{Kaushik Roy}.} \bibinfo{year}{2005}\natexlab{}.
\newblock \showarticletitle{Modeling of failure probability and statistical design of {SRAM array for yield enhancement in nanoscaled CMOS}}.
\newblock \bibinfo{journal}{\emph{{IEEE} Trans. Comput.-Aided Design Integr. Circuits Syst.}} \bibinfo{volume}{24}, \bibinfo{number}{12} (\bibinfo{year}{2005}), \bibinfo{pages}{1859--1880}.
\newblock


\bibitem[Pentapati et~al\mbox{.}(2021)]%
        {3DIC}
\bibfield{author}{\bibinfo{person}{Sai Surya~Kiran Pentapati}, \bibinfo{person}{Bon~Woong Ku}, {and} \bibinfo{person}{Sung~Kyu Lim}.} \bibinfo{year}{2021}\natexlab{}.
\newblock \showarticletitle{{ML-based wire RC prediction in monolithic 3D ICs} with an application to full-chip optimization}. In \bibinfo{booktitle}{\emph{Proc. ISPD}}. \bibinfo{pages}{75--82}.
\newblock


\bibitem[Ren et~al\mbox{.}(2020)]%
        {ParaGraph}
\bibfield{author}{\bibinfo{person}{Haoxing Ren}, \bibinfo{person}{George~F Kokai}, \bibinfo{person}{Walker~J Turner}, {and} \bibinfo{person}{Ting-Sheng Ku}.} \bibinfo{year}{2020}\natexlab{}.
\newblock \showarticletitle{{ParaGraph}: Layout parasitics and device parameter prediction using graph neural networks}. In \bibinfo{booktitle}{\emph{Proc. DAC}}. \bibinfo{pages}{1--6}.
\newblock


\bibitem[Schlichtkrull et~al\mbox{.}(2018)]%
        {rgcn}
\bibfield{author}{\bibinfo{person}{Michael Schlichtkrull}, \bibinfo{person}{Thomas~N Kipf}, \bibinfo{person}{Peter Bloem}, \bibinfo{person}{Rianne Van Den~Berg}, \bibinfo{person}{Ivan Titov}, {and} \bibinfo{person}{Max Welling}.} \bibinfo{year}{June 3--7, 2018}\natexlab{}.
\newblock \showarticletitle{Modeling relational data with graph convolutional networks}. In \bibinfo{booktitle}{\emph{Proc. European Semantic Web Conference (ESWC), Heraklion, Crete, Greece}}. \bibinfo{pages}{593--607}.
\newblock


\bibitem[Shen et~al\mbox{.}(2019)]%
        {lp2}
\bibfield{author}{\bibinfo{person}{Shan Shen}, \bibinfo{person}{Tianxiang Shao}, \bibinfo{person}{Xiaojing Shang}, \bibinfo{person}{Yichen Guo}, \bibinfo{person}{Ming Ling}, \bibinfo{person}{Jun Yang}, {and} \bibinfo{person}{Longxing Shi}.} \bibinfo{year}{2019}\natexlab{}.
\newblock \showarticletitle{{TS} cache: A fast cache with timing-speculation mechanism under low supply voltages}.
\newblock \bibinfo{journal}{\emph{{IEEE} Trans. {VLSI} Syst.}} \bibinfo{volume}{28}, \bibinfo{number}{1} (\bibinfo{year}{2019}), \bibinfo{pages}{252--262}.
\newblock


\bibitem[Shen et~al\mbox{.}(2023)]%
        {ultra8t}
\bibfield{author}{\bibinfo{person}{Shan Shen}, \bibinfo{person}{Hao Xu}, \bibinfo{person}{Yongliang Zhou}, \bibinfo{person}{Ming Ling}, {and} \bibinfo{person}{Wenjian Yu}.} \bibinfo{year}{2023}\natexlab{}.
\newblock \showarticletitle{Ultra8T: A Sub-Threshold 8T SRAM with Leakage Detection}.
\newblock \bibinfo{journal}{\emph{arXiv preprint arXiv:2306.08936}} (\bibinfo{year}{2023}).
\newblock


\bibitem[Shook et~al\mbox{.}(2020)]%
        {MLParest}
\bibfield{author}{\bibinfo{person}{Brett Shook}, \bibinfo{person}{Prateek Bhansali}, \bibinfo{person}{Chandramouli Kashyap}, \bibinfo{person}{Chirayu Amin}, {and} \bibinfo{person}{Siddhartha Joshi}.} \bibinfo{year}{2020}\natexlab{}.
\newblock \showarticletitle{{MLParest}: Machine learning based parasitic estimation for custom circuit design}. In \bibinfo{booktitle}{\emph{Proc. DAC}}. \bibinfo{pages}{1--6}.
\newblock


\bibitem[Veli{\v{c}}kovi{\'c} et~al\mbox{.}(2017)]%
        {gat}
\bibfield{author}{\bibinfo{person}{Petar Veli{\v{c}}kovi{\'c}}, \bibinfo{person}{Guillem Cucurull}, \bibinfo{person}{Arantxa Casanova}, \bibinfo{person}{Adriana Romero}, \bibinfo{person}{Pietro Lio}, {and} \bibinfo{person}{Yoshua Bengio}.} \bibinfo{year}{2017}\natexlab{}.
\newblock \showarticletitle{Graph attention networks}.
\newblock \bibinfo{journal}{\emph{arXiv preprint arXiv:1710.10903}} (\bibinfo{year}{2017}).
\newblock


\bibitem[Wang et~al\mbox{.}(2019)]%
        {dgl}
\bibfield{author}{\bibinfo{person}{Minjie Wang}, \bibinfo{person}{Da Zheng}, \bibinfo{person}{Zihao Ye}, \bibinfo{person}{Quan Gan}, \bibinfo{person}{Mufei Li}, \bibinfo{person}{Xiang Song}, \bibinfo{person}{Jinjing Zhou}, {et~al\mbox{.}}} \bibinfo{year}{2019}\natexlab{}.
\newblock \showarticletitle{Deep graph library: A graph-centric, highly-performant package for graph neural networks}.
\newblock \bibinfo{journal}{\emph{arXiv preprint arXiv:1909.01315}} (\bibinfo{year}{2019}).
\newblock


\bibitem[Wu et~al\mbox{.}(2020)]%
        {GNNsurvey}
\bibfield{author}{\bibinfo{person}{Zonghan Wu}, \bibinfo{person}{Shirui Pan}, \bibinfo{person}{Fengwen Chen}, \bibinfo{person}{Guodong Long}, \bibinfo{person}{Chengqi Zhang}, {and} \bibinfo{person}{S~Yu Philip}.} \bibinfo{year}{2020}\natexlab{}.
\newblock \showarticletitle{A comprehensive survey on graph neural networks}.
\newblock \bibinfo{journal}{\emph{{IEEE} Trans. Neural Netw. Learn. Syst.}} \bibinfo{volume}{32}, \bibinfo{number}{1} (\bibinfo{year}{2020}), \bibinfo{pages}{4--24}.
\newblock


\bibitem[Yang et~al\mbox{.}(2021)]%
        {CNNcap}
\bibfield{author}{\bibinfo{person}{Dingcheng Yang}, \bibinfo{person}{Wenjian Yu}, \bibinfo{person}{Yuanbo Guo}, {and} \bibinfo{person}{Wenjie Liang}.} \bibinfo{year}{2021}\natexlab{}.
\newblock \showarticletitle{{CNN-Cap}: Effective convolutional neural network based capacitance models for full-chip parasitic extraction}. In \bibinfo{booktitle}{\emph{Proc. ICCAD}}. \bibinfo{pages}{1--9}.
\newblock


\bibitem[Yang et~al\mbox{.}(2019)]%
        {cim1}
\bibfield{author}{\bibinfo{person}{Jun Yang}, \bibinfo{person}{Yuyao Kong}, \bibinfo{person}{Zhen Wang}, \bibinfo{person}{Yan Liu}, \bibinfo{person}{Bo Wang}, \bibinfo{person}{Shouyi Yin}, {and} \bibinfo{person}{Longxin Shi}.} \bibinfo{year}{2019}\natexlab{}.
\newblock \showarticletitle{{24.4 sandwich-RAM: An energy-efficient in-memory BWN architecture with pulse-width modulation}}. In \bibinfo{booktitle}{\emph{Proc. Int. Solid-State Circuits Conf. (ISSCC)}}. \bibinfo{pages}{394--396}.
\newblock


\bibitem[Yu et~al\mbox{.}(2022)]%
        {cim2}
\bibfield{author}{\bibinfo{person}{Chengshuo Yu}, \bibinfo{person}{Taegeun Yoo}, \bibinfo{person}{Kevin Tshun~Chuan Chai}, \bibinfo{person}{Tony Tae-Hyoung Kim}, {and} \bibinfo{person}{Bongjin Kim}.} \bibinfo{year}{2022}\natexlab{}.
\newblock \showarticletitle{{A 65-nm 8T SRAM compute-in-memory macro with column ADCs for processing neural networks}}.
\newblock \bibinfo{journal}{\emph{{IEEE} J. Solid-State Circuits}} \bibinfo{volume}{57}, \bibinfo{number}{11} (\bibinfo{year}{2022}), \bibinfo{pages}{3466--3476}.
\newblock


\end{thebibliography}
